\documentclass[conference,compsoc]{IEEEtran}

\ifCLASSOPTIONcompsoc
  \usepackage[nocompress]{cite}
\else
  \usepackage{cite}
\fi
\usepackage{color}
\ifCLASSINFOpdf
\else
\fi
\usepackage{algorithm2e}
\usepackage{algorithmic}
\usepackage{caption} 

\usepackage{multirow}
\usepackage{amsthm}
\usepackage{amsmath}
\usepackage{cases}
\usepackage[normalem]{ulem}
\usepackage{graphicx}
\usepackage[table,xcdraw]{xcolor}
\usepackage{color}

\begin{document}

\title{Mixed Strategy Game Model Against Data Poisoning Attacks}

\author{\IEEEauthorblockN{Yifan Ou}
\IEEEauthorblockA{McMaster University\\
Hamilton, Ontario, Canada\\
Email: ouyf@mcmaster.ca}

\and
\IEEEauthorblockN{Reza Samavi}
\IEEEauthorblockA{McMaster University, Vector Institute for Artificial Intelligence \\
Hamilton, Toronto, Ontario, Canada\\
Email: samavir@mcmaster.ca}
}

\maketitle

\begin{abstract}
In this paper we use game theory to model poisoning attack scenarios. We prove the non-existence of pure strategy Nash Equilibrium in the attacker and defender game. We then propose a mixed extension of our game model and an algorithm to approximate the Nash Equilibrium strategy for the defender. We then demonstrate the effectiveness of the mixed defence strategy generated by the algorithm, in an experiment.
\end{abstract}

\IEEEpeerreviewmaketitle

\section{Introduction}
Machine Learning (ML) algorithms are widely used in today's world. One of the major security threats to ML models is the poisoning attack, where an adversary tries to control a fraction of training data and inject malicious data points to degrade the model's performance or subvert the model outcome. Authors in \cite{Paudice2018} proposed an optimal poisoning attack, as well as a distance-based anomaly detection filtering mechanism for the defender. The success of this approach highly depends on selecting the correct filtering strength of its underlying detection mechanism. A highly-optimistic choice of filtering makes the ML model vulnerable as the attacker can craft an attack to inject extreme poisoned data points (far from centroid at the boundaries of feature space) and circumvent the detection model. An overly-pessimistic choice of filtering (removing points  relatively closer to centroid) strengthen the detection of poisoned data at the expense of ML model's accuracy due to the removal of genuine data points. 

An important problem to investigate is when the defence against poisoning attacks is expanded into a game \cite{Huang2011}, where the attacker aims to mislead the ML model while the defender deploys mechanism (e.g., filtering) to secure the algorithm. Studying behaviours of both parties when their strategies converge is a prerequisite of developing ML algorithms that are resistant to poisoning attacks while remaining maximally accurate. The objective of this paper is to find the Nash Equilibrium (NE) of the game model of poisoning attack and defense. Identifying the NE strategy will allow us to find the optimal filter strength of the defending algorithm, as well as the resulting impact to the ML model when both the attacker and the defender are using optimal strategies.

We are making three contributions: 1) we use game-theoretic modeling to formulate poisoning attacks, 2) we prove the non existence of pure NE in our game model, and 3) we propose a mixed NE strategy for the game model and show that the strategy is effective.

The rest of the paper is organized as follows. In Section \ref {relatedwork} we discuss a number of related work. In Section \ref {PNE} we formulate the poisoning attack and defence as a game model and provide the proof for non-existence of pure NE. In Section \ref{model} we propose the mixed strategy NE of our game model. In Section \ref {experiment} we report our preliminary experimental results and we conclude in Section \ref{conclusions}.

\section{Related Work}
\label {relatedwork}

\vspace{-.3cm}
Authors in \cite{Munoz2017} formulated a poisoning attack scenario as the following bi-level optimization problem and described efficient approaches to solve this problem:

\begin{equation*}
\begin{aligned}
& D_c \in \underset{D_c \in \phi(D)}{\text{argmax}}
\  O_A(D_{val}, w') \\
& \text{subject to}
 \ w' \in \underset{w' \in W}{\text{argmin}}
\  L(D_T \cup D_c, w')
\end{aligned}
\end{equation*}

\noindent
where $D_c$ is the set of malicious data points, $\phi(D)$ is a function that maps the malicious points onto the feasible domain, $O_A$ is the attacker's objective function, $D_{val}$ is the validation dataset used by the attacker, $w'$ is the trained classifier, and $D_T$ is the original training dataset, and $L$ is the loss function of the ML model. By solving this optimization problem, the attacker aims to find the best possible malicious data points to inject. Note that although the attacker may not have access to $D_T$ directly, he can acquire an auxiliary training dataset with a similar distribution to $D_T$, then the attacker can perform the attack to the auxiliary dataset and use the resulting set of malicious data points to contaminate $D_T$ \cite{Papernot2018T}.

To mitigate poisoning attacks, authors in \cite{Steinhardt2017} introduced distance-based filtering mechanisms for the defender, which is capable of mitigating the optimal poisoning attacks. To apply the filtering method, the defender computes the centroid for each class and removes the data points that are too far away from the centroid of each class. Authors in \cite{Paudice2018} proposed a similar distance based filtering mechanism by training an anomaly detection model using a \emph {trusted dataset} from the training data. There are other types of data sanitization techniques, such as \textit{Reject on Negative Impact(RONI)}\cite{Nelson2009} and PCA-based detection models\cite{Rubinstein2009}.

One limitation to the results in \cite{Paudice2018} and \cite{Munoz2017} is that, the attacker's strategy is chosen under the assumption that no defender is present. Furthermore, the defensive algorithm is shown to be effective only when the poisoning data points are placed optimally (and as a consequence, highly detectable). In reality, a sophisticated attacker would adjust his poisoning strategy, taking into account the defensive mechanism, while the defender is also updating his strategy accordingly. In that situation, we can view the poisoning attack as a competitive game between the attacker and the defender. Authors in \cite{Steinhardt2017} investigated the attacker's strategy as a \emph {minimax game} with the defender, but they only considered the case in which the defender uses a \emph{pure strategy} (i.e. the strength of filter is fixed and is known by the attacker). In this case, the obvious optimal attack strategy is to place the poisoning points close to the boundary of the filter. In our research we omit the assumption of \emph{pure strategy} and in fact we prove its non-optimality. Studying the Nash Equilibrium(NE) of this game model is highly important as it not only provides us the insight to a decent defense strategy, but also allows us to estimate the accuracy impact on the ML model under different poisoning attacks.

\vspace{-.3cm}
\section{Proof of Nash Equilibrium}
\label {PNE}
\vspace{-.3cm}
In this section we first provide three preliminary definitions of Game theories, Nash equilibrium and Zero-sum games. Then we provide a proof for non-existence of pure-strategy in the attacker-defender game model in poisoning attack scenarios.

\vspace{.1cm}
\noindent
\textbf{Game Theories}:
The poisoning attack scenario can be formulated as a two player competitive game model. In this game model, $X_1$ and $X_2$ denote as the pure strategy sets (the set of legal moves) of player 1 and player 2 respectively, $n_1$ and $n_2$ represents the cardinality of $X_1$ and $X_2$. Let $X = X_1 \times X_2$ represents all possible combinations of choices made by the players with cardinality $n_1 * n_2$. Then each player's payoff function is denoted as $u_i: X \rightarrow{\mathbf{R}}$, and the global payoff function will be denoted as $U(x) = \{u_1(x),u_2(x)\}$ for every $x \in X$.

\vspace{.1cm}
\noindent
\textbf{Nash Equilibrium}: Nash Equilibrium (NE) is the equilibrium strategies of all players in a competitive game. When all players in the game are using the NE strategies, no players can benefit by changing his strategy. According to Nash's Existence Theorem \cite{Nash48}, every finite game has at least one NE. For infinite games, where each player's choice is from an infinite set (e.g., choosing a price for his product, which could be any real number), NE also exists if the set of choices are compact and the payoff function is continuous \cite{Glicksberg1952}. An NE can be a \emph {pure strategy} equilibrium, where all players will stick with one choice all the time (e.g., in a game where each player picks a number simultaneously from 0 to 10, the highest number wins; thus all players will pick 10 all the time). However, a game will more likely have a mixed strategy NE, where each player's strategy is a probability distribution over all possible choices (e.g., in Rock-Paper-Scissors, the NE strategy is to pick rock, paper and scissors 1/3 of the times). 

 \vspace{.1cm}
 \noindent
\textbf{Zero-sum games}: In zero-sum games, the total benefit to all players in the game, for every combination of strategies, always adds to zero. In other words, a player benefits only at the equal expense of others. More formally, the players' utility function holds at $u_1(x) = -u_2(x)$. According to \cite{Reny1999}, a game is payoff secure if for every $x \in X$ and every $\epsilon > 0$, each player can secure a payoff of $U(X) - \epsilon$ at x. That is, each player has a way to guarantee his payoff at every $x \in X$, even if the other player deviates from $x$ slightly. All continuous games (infinite games with continuous payoff functions) possesses this property, but many discontinuous games also do, as described in \ref{MixedNE}.

\vspace{-.2cm}
\subsection{Non-Existence of Pure Strategy NE}
\vspace{-.3cm}
In our game model, the attacker
(denoted by $a$) chooses a set $S_a = \{[r_1, n_1], [r_2, n_2], ... [r_m, n_m]\}$, where for each radius $r_i \in S_a$, $n_i$ poisoning points will be placed optimally within $r_i$ distance from the centroid of the original dataset ($\sum_{i=1}^m n_i =N$, where $N$ is the total number of maliciously injected points). Since the poisoning points are placed optimally, we can expect their locations to be near the boundary of the hypersphere with radius $r_i$. The defender (denoted by $d$) also chooses $\theta_d$ as the radius of the filter. Any data points outside the hypersphere centered at the centroid of the \emph{original} dataset with radius $\theta_d$ will be removed. This strategy by the defender is justified even though the defender does not have access to the original dataset, and the attacker's poisoned points may influence the position of the centroid. This is because the influence is dependant on the method used by the defender to compute the centroid (e.g., using median vs. mean),  as long as the defender uses a \emph{good} method to find the centroid (i.e. a method less affected by the outliers), and the proportion of malicious points in the training dataset is relatively small, the position of the centroid will not be changed drastically by the malicious datapoints. In this game, the attacker and the defender choose $S_a$ and $\theta_d$ simultaneously, and the outcome of the game is determined as the followings:
\begin{itemize}
    \item If a poisoning point $i$ is not removed (i.e. $\theta_d \geq r_i \in S_a$), the attack produces $E(r_i, n_i)$ payoff for the attacker. The greater $r_i$ is, the higher the payoff for the attacker. The attack yields $\sum_{\{r_i \in S_a | \theta_d \geq r_i\}} E(r_i, n_i)$ payoff for the attacker in total.
    \item In addition, the defender pays additional cost $\Gamma(\theta_d)$ for removing genuine data points. The smaller $\theta_d$ is, the higher the cost.
\end{itemize}

  \noindent
  Thus the payoff functions can be represented as: $U(S_a, \theta_d) = U_a(S_a, \theta_d) = - U_d (S_a, \theta_d) = \sum_{\{r_i \in S_a | \theta_d \geq r_i\}} E(r_i, n_i) + \Gamma(\theta_d)$

 \vspace{.2cm}
 \noindent
  \textbf{Proposition 1}: In this game model, a pure strategy Nash Equilibrium is very unlikely to exist. 
  
 \noindent
\begin{proof} The best respond functions (BRF)\cite{Nash48} of the attacker and the defender are:

\vspace{-.2cm}
\begin{subnumcases}{\beta_a(\theta_d) = }
    {[\theta_d, N]} & $\theta_d \geq T_a$ \label{ordinaryA}
    \\
    \text{all }  r_i \geq T_a & otherwise \label{blockedA}
\end{subnumcases}

 \begin{subnumcases}{\beta_d(S_a) =}
    B & if $r_i \leq T_d, \forall r_i \in S_a$ \label{blockedD}
    \\
     r_i - \epsilon & $r_i = \min \{r_i \in S_a | r_i > T_d\}$ \label{ordinaryD}

\end{subnumcases}

where, 
\begin{itemize}
    \item $T_a$ is the minimum distance from the centroid where poisoning points yield benefit for the attacker: $E(r_i, n_i) \leq 0, \forall r_i <= T_a$. Since $E(r_i, n_i)$ is not affected by $\theta_d$ (only $\sum_{\{r_i \in S_a | \theta_d \geq r_i\}} E(r_i, n_i)$ is affected by $\theta_d$), the value of $T_a$ is only dependant on the distribution of the \emph{original} dataset.
    \item $T_d$ is the distance from the centroid where moving $\theta_d$ towards the centroid causes the defender's payoff to be strictly lower than $U_d(S_a, T_d)$. $U_d(S_a, i) < U_d(S_a, T_d), \forall i < T_d$. This value is determined by the \emph{original} dataset and the attacker's poisoning strategy (this is not known by the defender in practice, but is valid under BRF analysis).
    \item $B$ is the boundary, aka maximum possible distance from the centroid.
\end{itemize}
  Suppose that $(S_{a}^*, \theta_d^*)$ are the NE strategies. Then 
  \[
  \begin{cases}
  (S_{a}^*, \theta_d^*) \in \beta_d(S_{a}^*) \\
   (\theta_d^*, S_{a}^*) \in \beta_a(\theta_d^*)
  \end{cases}
  \]
In such condition, the NE strategy set $(S_{a}^*, \theta_d^*)$ must satisfy one of \ref{ordinaryA} and \ref{blockedA}, as well as one of \ref{ordinaryD} and \ref{blockedD}. Denote $r_{min}$ as the minimum radius in $S_a$ such that $r_{min} > T_d$. In the remainder of the proof, we relax the condition above into the following: 

  \[
  \begin{cases}
  (r_{min}^*, \theta_d^*) \in \beta_{d}(S_a^*) \\
   (\theta_d^*, r_{min}^*) \in \beta_{a}(\theta_d^*)
  \end{cases}
  \]

\noindent
Note that each choice of $r_{min} \in \beta_{a}(\theta_d)$ may represent multiple possible attacker's responses $S_a$. Therefore, if the NE ($r_{min}^*, \theta_d^*$) does not exist for this relaxed problem, the NE for the original problem will not exist either.
Clearly \ref{ordinaryA} and \ref{ordinaryD} cannot be satisfied simultaneously, as $\beta_{a}(\theta_d) = r_{min}$ and $\beta_{d}(r_{min}) = r_{min} - \epsilon$ does not intersect. Similarly \ref{blockedA} and \ref{blockedD} does not intersect because \ref{blockedA} has condition $\theta_d < T_a$, but $\beta_{d}(S_a) = B$ and $B >> T_a$.  \ref{ordinaryA} and \ref{blockedD} also does not intersect, as $\beta_{a}(\theta_d) = r_{min}$ intersects $\beta_{d}(r_{min}) = B$ at $(B, B)$, which violates the condition of \ref{blockedD}.

The only possible intersection of the BRFs is \ref{blockedA} and \ref{ordinaryD}. This will occur when $T_a \geq T_d$, at $(\theta_d = T_a, r_{min} = T_a + \epsilon)$. However, it is impossible to have $T_a > T_d$, because the attacker will not place poisoning points inside $T_a$ (doing so yields no profit). If the defender move $\theta_d$ from $T_a$ towards $T_d$, he will lose from $\Gamma(\theta_d)$ and gain nothing from $E(S_a)$, thus violates the definition of $T_d$. In very rare cases, $T_a$ may be exactly equal to $T_d$, but such situation is very uncommon and therefore its NE is inapplicable to most ML applications. 
\end{proof}

\vspace{-.3cm}
\section {Poisoning Attack Threat Model}
\label{model}
\vspace{-.3cm}
Given non-existence of \emph {pure strategy} NE, in this section we first prove that a \emph {mixed strategy} NE exists and then propose the \textit{mixed extension} of our game model. 

\vspace{-.3cm}
\subsection{Mixed Strategy NE} \label{MixedNE}

\vspace{-.3cm}
When players adapting mixed strategies, the defender does not have to use the same value of $\theta_d$ all the time. Instead, he may have a set of possible values for $\theta_d$ and choose the value based on a probability distribution in each game. Similarly, the attacker can choose his strategy using a probability distribution. Denote $M_d$ as the set of probability measure on the Borel subset of all possible values of $\theta_d$, $M_a$ as the probability measure on the Borel subset of all $S_a$. $M_a$ and $M_d$ are the \textit{mixed strategy sets}. Denote $X_a$ and $X_d$ as the \textit{pure strategy set} of the attacker and the defender respectively, and $X = X_a \times X_d$, then we can extend the utility function into $U(\sigma \in (M_a \times M_d) ) = \int_{X} U(S_a, \theta_d) d\sigma$, where $\sigma$ represents the mixed strategy chosen by the attacker and the defender. 

 \vspace{.2cm}
 \noindent
  \textbf{Proposition 2}: In this game model, a mixed strategy NE exists. 
  
\vspace{-.2cm}
\begin{proof} 
According to \cite{Reny1999}, if the mixed-extension of the game is \textit{reciprocally upper semi-continuous} and \textit{payoff secure}, then it has a mixed strategy NE. 
A zero-sum game is always reciprocally upper semi-continuous in its \textit{mixed extension} \cite{Reny1999}. Furthermore, our game model is payoff secure, because decreasing $\theta_d$ slightly will at worst decrease $U_d$ slightly, if we make a continuous approximation on $\Gamma(\theta_d)$, as long as the attacker's mixed strategy does not change too much. Similarly, decreasing a radius in $S_a$ slightly will at worst decrease $U_a$ slightly. Therefore, our game possesses a mixed strategy NE. 
  \end{proof}

\vspace{-.3cm}
\subsection{Game Model} \label{GameModel}
\vspace{-.3cm}
For this research, we will model the poisoning attack scenario as a competitive zero-sum game \cite{bin2007} between the attacker and the defender, who uses the filtering mechanism described in \cite{Paudice2018}. In our game model, the attacker starts by injecting malicious data set $D_C$ into the original training data $D_T$. Then, the defender calculates the radius of the filter $\theta$ using the estimated percentage of malicious data, and applies the filtering algorithm $A(D_T,\theta) \rightarrow D_F \subseteq D_T$. Practically, the attacker will not know $\theta$ ahead of time, and the defender is unaware of the attack. Therefore we can safely model their behaviours as choosing $D_C$ and $\theta$ simultaneously, then evaluate the outcome using the attacker’s objective function $O_a (D_{val},D_C,\theta)$, where $D_{val}$ is the data set used to evaluate the attacker’s objective. The distribution of $D_{val}$ is solely dependant on the attacker’s goal, while $D_C$ and $\theta$ are used to train the model. We can clearly see that this is a continuous competitive game model, as the attacker is trying to maximize the value of $O_a (D_{val},D_C,\theta)$, while the defender aims to minimize it.
 
 One of the key properties of the NE of this game model is that, the defender's strategy must meet the following conditions:
 \begin{enumerate}
     
     \item The defender uses a mixed strategy $m \in M_d$ with at least two $pdf(\theta_d) > 0$, where $pdf$ is the probability density function of $m$. \label{ms}
     \item For every $\theta_d$ in the defender's mixed strategy $m \in M_d$ with $pdf_m(\theta_d) > 0$, the product of $E(\theta_d)$ and $cdf_{m}(\theta_d)$ must be equal, where $cdf$ is the cumulative density function of $m$, counting from $B$ towards the centroid.\label{cdf}
 \end{enumerate}
 \begin{proof} Suppose that a defender's strategy $m^*$ is a NE strategy that does not meet one of those conditions. In the previous section we have shown that condition \ref{ms} must be satisfied for all NE strategies in our game model. Although the attacker is allowed to use mixed strategy, if the defender uses only pure strategy, then the attacker's best response is also in pure strategy. Therefore neither the attacker nor the defender need to adjust their BRF, hence condition \ref{ms} can be validated with the proof procedure in the previous section.
 If condition \ref{cdf} is not satisfied, then: 
 
\vspace{.1cm}
$\exists (\theta_x, \theta_y) \in m^* | cdf_{m^*}(\theta_x) * E(\theta_x) > cdf_{m^*}(\theta_y) * E(\theta_y) \wedge pdf_{m^*}(\theta_x) > 0,  pdf_{m^*}(\theta_y) > 0$ \\

\vspace{-.2cm} 
 Then, the attacker will not place any malicious points on $\theta_y$, because doing so yields less profit than placing on $\theta_x$. 
 Then, the defender may increase the value of $\theta_y$ until $cdf_{m^*}(\theta_x) * E(\theta_x) = cdf_{m^*}(\theta_y) * E(\theta_y)$ (or if $\theta_y = B$, we can shift probability from $\theta_x$ to $\theta_y$). This will reduce $\Gamma(m)$ and hence increase the defender's profit without altering the attacker's choice of strategy ($\theta_x$ remains no less attractive than $\theta_y$ throughout this process). This contradicts with the assumption that $m^*$ is the NE strategy. 

 We have shown that any strategy that does not meet the above conditions are not NE strategies, therefore a NE strategy must possesses the properties above.
 \end{proof}
 
 Note that when these conditions are met, malicious points placed on each $\theta_d$ with $pdf_m(\theta_d) > 0$ yields the same profit for the attacker. Therefore the attacker is \emph{indifferent} among his strategies (as long as all malicious points are placed on some $\theta_d$ with $pdf_m(\theta_d) > 0$ in any combination). The NE strategy of the defender is simply the strategy which minimizes the attacker's profit while satisfying the conditions above. Using these properties, we can derive an algorithm to approximate the defender's NE strategy (Algorithm \ref{alg:computeDefense}). The algorithm starts with a set of initial filter radii. In every iteration, the algorithm calculates the probability that satisfies the conditions described in the proof above for every radius in the set. Then it performs gradient descent on the set of radius to minimize the defender's loss function. Notice that placing all poisoning points within the strongest filtering radius $r_{min}$ is one of the optimal attack strategy, therefore we can safely use its resulting loss $N * E(r_{min})$ to represent the loss from an optimal attack.
 
 Computing an exact NE strategy may be time consuming and infeasible due to the unbounded number of radius that the defender can include in his mixed strategy. However, computing the NE strategy which uses a fixed number of radius is possible and is usually sufficient in practice as described in the next section.

\begin{algorithm}
 \LinesNumbered

        \textbf{INPUT:} 
        \begin {enumerate}
        \item $\Gamma(p)$ - the estimated loss for removing genuine points(p = fraction of points to remove)\\ 
        \item $E(p)$ - the maximum effect of a poisoning point placed in that percentile\\
        \item $n$ - number of radius in mixed strategy\\
        \item $\epsilon$ - convergence threshold\\
        \item $N$ - expected number of poisoning data points in the dataset
        \end {enumerate}
        \textbf{OUTPUT:} 
        \begin {enumerate}
        \item $M_d$ - the NE mixed strategy of defender\\
        \item $U_d(M_d, *)$ - the resulting impact to the ML model
        \end {enumerate}

            $\{r_1, r_2, ..., r_n\}$ = chooseInitialRadius($n$)
            $S_r = \{r_1, r_2, ..., r_n\}$\\
            $t = 0$ \\
            \While {$ f(S_r)^t - f(S_r)^{t-1} < \epsilon$}{
                $pdf$ = findPercentage($S_r$)\\ 
                $r_{min} = $ min$(S_r)$
                $f = N * E(r_{min}) + \int_0^1 pdf(p_i) * \Gamma(p_i) dp_i$\\
                Compute $\nabla(f(S_r)) = \frac{df}{dS_r}$\\
                $S_r = S_r - \nabla(f(S_r))$\\
                $t = t + 1$
            }
            \textbf{return} $\{S_r, pdf\}$, $f(S_r)$
       \linebreak
       \caption{Compute Optimal Defense}
        \label{alg:computeDefense}
\end{algorithm}
\vspace{-.3cm}
 
\vspace{-.3cm}
\section{Experiments}
\label{experiment}
\vspace{-.3cm}
The goal of our experiment is to assess the effectiveness of our mixed-strategy defense in terms of the changes in the accuracy of the algorithm when optimal poisoning attacks are performed. The accuracy of algorithm without our mixed strategy is considered as the benchmark for comparison. We used the \textit{Spambase} dataset\footnote{https://archive.ics.uci.edu/ml/machine-learning-databases/spambase/spambase.data} to train and test our ML model.

\begin{figure}[t]
        \center{\includegraphics[width=70mm]
        {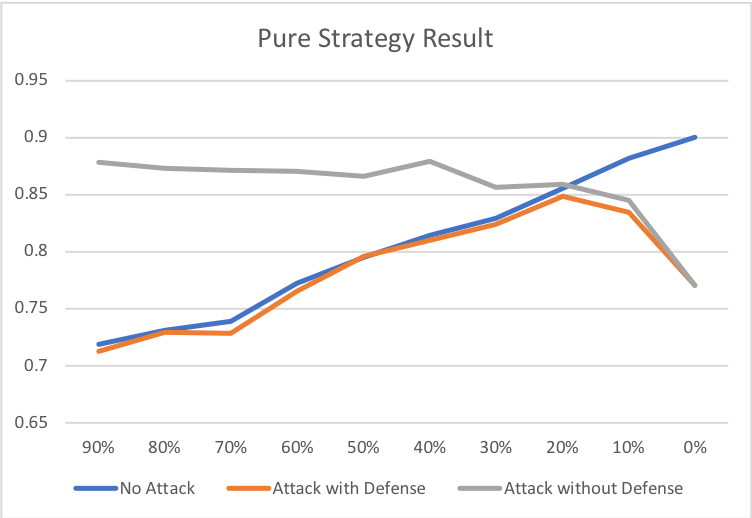}}
        \caption{\label{fig:pure_result} Pure strategy defense under optimal attack }
      \end{figure}

To setup the experiment, we started by assessing the accuracy impact to the ML model under pure strategy attack and defense strategy. Using pure defense strategy, we made the assumption that the attacker has full knowledge to the ML model, including the defense strategy itself. Therefore the \textit{optimal} attack in this scenario is to place all poisoning points near the boundary of the filter. We start by loading the Spambase dataset, separating the 4601 instances into 70\% of training data (3220 instances) and 30\% of test data (1381 instances). Then we performed \textit{optimal} poisoning attack and filtering on the \textit{training} dataset. We assumed that the attacker can manipulate 20\% of the training data.  We used Support Vector Machine (SVM) with hinge loss as our ML model and trained it for 5000 epoch in every iteration. We tested the model's accuracy with different filtering strength, and the results are reported in Fig.  \ref{fig:pure_result}. The y-axis is the accuracy of the ML model, and the x-axis represents the percentage of data points removed by the filter, which is the same as 1-percentile of poisoning data. 

As shown in this figure, applying the filter reduces the accuracy of the ML model, regardless of the presence of the attack. However, aborting the filter will enable the attacker to perform more threatening attacks. Also, we can see that for this model, the defender loses incentive to increase filter strength at some point between 10\% and 30\%, while the attacker always have incentive to inject, regardless of the percentile. These facts indicate that no pure strategy NE exists in this model. 

We then ran Algorithm~\ref{alg:computeDefense} to generate the mixed defense strategy. The input of the algorithm, $E(p)$ and $\Gamma(p)$, are approximated using the results in Fig. \ref{fig:pure_result}. We tested the accuracy of the ML model when the generated defense strategy is used. Note that the \textit{optimal} attack in this case is to place poisoning points near \textit{any} boundary of the mixed defense strategy in any combination, as explained in section \ref{GameModel}. The results of the mixed strategy defence are reported in Table \ref{tab:mixed_result}. The number of radius is the input to the algorithm \ref{alg:computeDefense},  the radii and probabilities are the outputs of the algorithm. 

As shown in this table, the accuracy of the ML model using mixed defense strategy is strictly higher than the accuracy of all pure defense strategies. This validates the effectiveness of mixed strategy defense in poisoning attacks. We experimented filters with $n \leq 5$, the accuracy of the resulting model stays roughly the same after $n = 3$. The defender's strategy becomes a closer approximation to NE as the value of $n$ increases. However, the computation time increases significantly when computing high value of $n$.

\begin{table}[]
\begin{tabular}{|c|c|c|c|c|c|}
\hline
\# radius   & \multicolumn{2}{c|}{2}      & \multicolumn{3}{c|}{3}                                                                        \\ \hline
Radius      & 5.8\%         & 15.7\%      & 5.8\%  & 9.4\%  & 16.3\% \\ \hline
Probability & 51.2\%        & 48.8\%      & 33.3\% & 33.3\% &  33.4\% \\ \hline
Accuracy    & \multicolumn{2}{c|}{85.6\%} & \multicolumn{3}{c|}{86.1\%}                                            \\ \hline
\end{tabular}

\caption{\label{tab:mixed_result} Mixed strategy defense under optimal attack}
\end{table}

\section {Conclusions}
\label {conclusions}
In this paper, we used game theory to model the attacker and defender strategies in poisoning attack scenario. We proved non-existence of the pure strategy Nash Equilibrium, proposed a mixed extension of our game model and an algorithm to approximate the Nash Equilibrium strategy for the defender, then demonstrated the effectiveness of the mixed defence strategy generated by the algorithm.
One limitation to our research is that, we used the results from the pure strategy scenario to approximate $E(p)$ and $\Gamma(p)$ in order to compute the optimal mixed defense strategy. It is possible that a generalized $E(p)$ and $\Gamma(p)$  exists across all datasets, which we leave it for future research. Another interesting, yet more general approach, to address poisoning attacks which we are interested to investigate is based on detecting and rejecting samples using auditing algorithms (see \cite{Samavi2017ashort} as one general approach for auditing). This approach is particularly useful when the users' feedback are sought in an online fashion to update and improve the trained model. 

\section*{Acknowledgment}
Supports from NSERC Canada and Vector Institute for Artificial Intelligence are acknowledged.

\ifCLASSOPTIONcompsoc

\else

\fi

\bibliography{main}

\begin{thebibliography}{10}

\bibitem{Paudice2018}
A.~Paudice, L.~Muñoz-González, A.~Gyorgy, and E.~C. Lupu, ``Detection of
  adversarial training examples in poisoning attacks through anomaly
  detection,'' {\em arXiv:1802.03041}, 2018.

\bibitem{Huang2011}
L.~Huang, A.~D. Joseph, B.~Nelson, B.~I. Rubinstein, and J.~D. Tygar,
  ``Adversarial machine learning,'' in {\em Proceedings of the 4th ACM Workshop
  on Security and Artificial Intelligence}, AISec '11, (New York, NY, USA),
  pp.~43--58, ACM, 2011.

\bibitem{Munoz2017}
L.~Mu\~{n}oz Gonz\'{a}lez, B.~Biggio, A.~Demontis, A.~Paudice, V.~Wongrassamee,
  E.~C. Lupu, and F.~Roli, ``Towards poisoning of deep learning algorithms with
  back-gradient optimization,'' in {\em Proceedings of the 10th ACM Workshop on
  Artificial Intelligence and Security}, AISec '17, (New York, NY, USA),
  pp.~27--38, ACM, 2017.

\bibitem{Papernot2018T}
N.~Papernot, P.~D. McDaniel, and I.~J. Goodfellow, ``Transferability in machine
  learning: from phenomena to black-box attacks using adversarial samples,''
  {\em CoRR}, vol.~abs/1605.07277, 2016.

\bibitem{Steinhardt2017}
J.~Steinhardt, P.~W.~W. Koh, and P.~S. Liang, ``Certified defenses for data
  poisoning attacks,'' in {\em Advances in Neural Information Processing
  Systems 30} (I.~Guyon, U.~V. Luxburg, S.~Bengio, H.~Wallach, R.~Fergus,
  S.~Vishwanathan, and R.~Garnett, eds.), pp.~3517--3529, Curran Associates,
  Inc., 2017.

\bibitem{Nelson2009}
B.~Nelson, M.~Barreno, F.~Jack~Chi, A.~D. Joseph, B.~I.~P. Rubinstein,
  U.~Saini, C.~Sutton, J.~D. Tygar, and K.~Xia, {\em Misleading Learners:
  Co-opting Your Spam Filter}, pp.~17--51.
\newblock Boston, MA: Springer US, 2009.

\bibitem{Rubinstein2009}
B.~I. Rubinstein, B.~Nelson, L.~Huang, A.~D. Joseph, S.-h. Lau, S.~Rao,
  N.~Taft, and J.~D. Tygar, ``Antidote: understanding and defending against
  poisoning of anomaly detectors,'' in {\em Proceedings of the 9th ACM SIGCOMM
  conference on Internet measurement}, pp.~1--14, ACM, 2009.

\bibitem{Nash48}
J.~F. Nash, ``Equilibrium points in n-person games,'' {\em Proceedings of the
  National Academy of Sciences}, vol.~36, no.~1, pp.~48--49, 1950.

\bibitem{Glicksberg1952}
I.~L. Glicksberg, ``A further generalization of the kakutani fixed point
  theorem, with application to nash equilibrium points,'' {\em Proceedings of
  the American Mathematical Society}, vol.~3, no.~1, pp.~170--174, 1952.

\bibitem{Reny1999}
P.~J. Reny, ``On the existence of pure and mixed strategy nash equilibria in
  discontinuous games,'' {\em Econometrica}, vol.~67, no.~5, pp.~1029--1056,
  1999.

\bibitem{bin2007}
K.~Binmore, {\em Playing for Real: A Text on Game Theory}.
\newblock Oxford University Press, 2007.

\bibitem{Samavi2017ashort}
R.~Samavi and M.~P. Consens, ``{Publishing Privacy Logs to Facilitate
  Transparency and Accountability},'' {\em Journal of Web Semantics}, vol.~50,
  pp.~1--20, 2018.

\end{thebibliography}
\bibliographystyle{ieeetr}

\end{document}